\documentclass{article} 
\usepackage[final]{colm2026_conference}
\usepackage{textcomp}
\usepackage{microtype}
\usepackage{hyperref}
\usepackage{url}
\usepackage{booktabs}
\usepackage{multirow} 
\usepackage{graphicx}
\usepackage{booktabs}
\usepackage{subcaption} 
\usepackage[normalem]{ulem}
\usepackage{amsmath} 
\usepackage{svg}
\usepackage{paralist}
\usepackage{colortbl}
\usepackage{algorithm}
\usepackage{algpseudocode}
\usepackage{hyperref}
\usepackage{caption} 
\usepackage{subcaption} 

\usepackage[dvipsnames]{xcolor}   
\usepackage{amssymb}  

\usepackage{lineno}

\definecolor{darkblue}{rgb}{0, 0, 0.5}
\hypersetup{colorlinks=true, citecolor=darkblue, linkcolor=darkblue, urlcolor=darkblue}

\title{Only Ask What You Don't Know: Grounded Delta Planning for Efficient Multi-step RAG}

\author{Wei-Chieh Chou\textsuperscript{1}\thanks{Contributed equally as first authors.} \quad 
  Xuanjun Chen\textsuperscript{2}\footnotemark[1] \quad 
  Jian-Ren Lin\textsuperscript{3}\thanks{Contributed equally as second authors.} \quad 
  Claire Lin\textsuperscript{4}\footnotemark[2] \\
\textbf{Hung-yi Lee}\textsuperscript{ 2, 5}, \textbf{Jyh-Shing Roger Jang}\textsuperscript{ 1}\\
  \textsuperscript{1}Dept. of Computer Science and Information Engineering, National Taiwan University \\
  \textsuperscript{2}Graduate Institute of Communication Engineering, National Taiwan University \\
  \textsuperscript{3}Department of Economics, National Taiwan University \\
  \textsuperscript{4}Department of Information Management, National Taiwan University \\
  \textsuperscript{5}NTU Artificial Intelligence Center of Research Excellence (NTU AI-CoRE) 
}

\begin{document}

\ifcolmsubmission
\linenumbers
\fi

\maketitle

\begin{abstract}
    Multi-hop question answering remains challenging for Retrieval-Augmented Generation (RAG) because existing approaches either propagate errors across iterative retrieval rounds or over-generate reasoning steps, increasing cost without improving accuracy. We propose Grounded Delta Planning RAG (GDP-RAG), a plan-based framework that targets only the information delta ($\Delta$) based on three simple design choices: (1) preliminary retrieval to ground planning before execution, (2) a gap-conditioned planning prompt that asks only for missing information, and (3) a skeletal trajectory that pairs each subquery with a \texttt{Thought} capturing evidence from preliminary retrieval and carrying it through to the final answer. GDP-RAG focuses computation on unresolved gaps, yielding concise, reliable reasoning trajectories. Extensive experiments on HotpotQA, 2WikiMultiHopQA, and MuSiQue show that GDP-RAG achieves the highest accuracy (60.63\%) among all compared systems while maintaining a cost-of-pass of 0.51, 22\% lower than PAR-RAG (0.65) and 68\% lower than KnowTrace (1.57), with no method achieving both higher accuracy and lower cost.
\end{abstract}

\section{Introduction}
Large Language Models (LLMs) exhibit strong reasoning capabilities \citep{kojima2022zeroshot} but remain limited in knowledge-intensive scenarios, where reliance on parametric knowledge often leads to hallucinations.
Retrieval-Augmented Generation (RAG) \citep{lewis2020rag} alleviates this issue by grounding generation in external knowledge; however, it becomes insufficient for complex information needs such as multi-hop question answering \citep{talmor2018complexwebq, welbl2018qangaroo, lin2025ragtaiwan}, which require coordinated reasoning over distributed context.
Consequently, modern RAG systems increasingly adopt question decomposition to integrate retrieval with reasoning \citep{gao2023ragsurvey, singh2025agenticrag}. 

Existing approaches can be categorized by when decomposition is performed: step-wise and plan-based frameworks.  
Step-wise methods, such as ReAct \citep{yao2023react} and Self-Ask \citep{press2023compositionality}, dynamically interleave retrieval and reasoning by decomposing queries at each step.  
While flexible, this paradigm suffers from two critical limitations:  
(i) the accumulation of retrieved documents can exceed the model’s effective context capacity, leading to the lost-in-the-middle problem \citep{liu2024lost}; and  
(ii) strong inter-step dependencies amplify error propagation, whereby early mistakes cascade through the entire reasoning process \citep{dziri2023faith}.
In contrast, plan-based frameworks such as ReWOO \citep{xu2023rewoo} constructing static global plans before the execution. This design alleviates context overflow and reduces cascading errors by isolating sub-steps in advance, providing structural stability.  
However, this poses the risk of over or under-planning. Furthermore, strictly following such rigid plans leads to severe efficiency issues as the cost of each execution step becomes increasingly substantial \citep{singh2025agenticrag}.

\begin{figure*}[t]
    \centering
    \includegraphics[width=\textwidth]{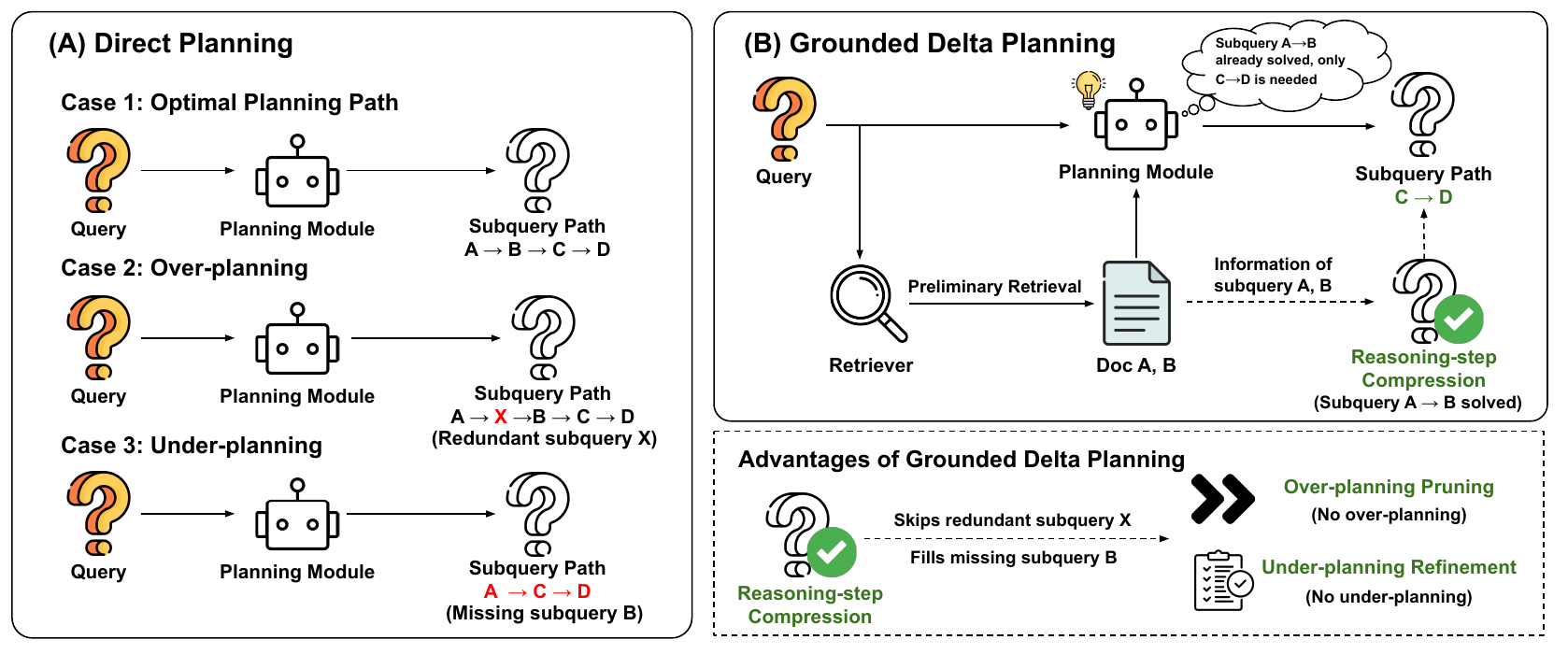}
    \caption{(A) Direct Planning decomposes queries using only parametric knowledge, producing      
  redundant subqueries (over-planning, e.g., X in Case 2) or missing ones (under-planning,
   e.g., B in Case 3). (B) Grounded Delta Planning retrieves relevant documents first,    
  then generates subqueries only for missing information ($\Delta$), enabling:
  Reasoning-step Compression that skips resolved steps (Case 1), Over-planning Pruning    
  that eliminates X (Case 2), and Under-planning Refinement that recovers B and preserves
  it in the \texttt{Thought} field (Case 3).}
    \label{fig:decomposition_comparison}
\end{figure*}

To address these challenges, we propose \textbf{Grounded Delta Planning RAG (GDP-RAG)}, a plan-based framework that plans only for the information delta ($\Delta$), the knowledge still missing after an preliminary retrieval.
GDP-RAG realises this idea through three design choices:
(1) preliminary retrieval to ground the planner in available evidence before decomposition;
(2) a gap-aware query decomposition that instructs the planner to identify what is already known and generate sub-questions only for remaining gaps; and
(3) a skeletal trajectory that pairs each subquery with a \texttt{Thought} capturing evidence from preliminary retrieval and carrying it through to the final answer.
Together, these choices yield compact plans that target genuine information gaps, avoiding both the over-planning of conventional plan-based methods and the error propagation of step-wise approaches.
As illustrated in Figure~\ref{fig:decomposition_comparison}, Direct Planning decomposes queries using only parametric knowledge, producing redundant or missing subqueries. Grounded Delta Planning conditions the planner on preliminary retrieval, enabling three structural advantages: Reasoning-step Compression that skips steps already resolved by retrieved context, Over-planning Pruning that eliminates redundant subqueries, and Under-planning Refinement that recovers missing dependencies through broader retrieval semantics. Our contributions are summarized as follows:
\begin{itemize}
\item We propose GDP-RAG, a plan-based multi-step RAG framework built on Grounded Delta Planning that retrieves first and plans only for the information delta, yielding concise reasoning trajectories without sacrificing answer quality.
\item Experiments on HotpotQA, 2WikiMultiHopQA, and MuSiQue show that GDP-RAG achieves the highest accuracy (60.63\%) among compared systems while maintaining a cost-of-pass of 0.51, 22\% lower than PAR-RAG (0.65) and 68\% lower than KnowTrace (1.57), with no method achieving both higher accuracy and lower cost.
\item Ablation and effectiveness analyses confirm that each design choice contributes, and that Grounded Delta Planning compresses reasoning steps, prunes redundant sub-questions, and refines incomplete plans.
\end{itemize}

\section{Related Work}
We categorize prior work on complex QA into
(1) Step-wise Approaches,
(2) Plan-based Approaches,
(3) Retrieval and Verification, and
(4) Improving the Planner and Reasoner.

\subsection{Step-wise Approaches}
Step-wise methods interleave retrieval and reasoning, generating one subquery at a time conditioned on previously retrieved context.
ReAct \citep{yao2023react}, IRCoT \citep{trivedi2023ircot}, and Iter-RetGen \citep{shao2023iterretgen} integrate retrieval into Chain-of-Thought reasoning \citep{wei2022chain}.
RAT \citep{wang2024rat} and Verify-and-Edit \citep{zhao2023verify} further refine reasoning chains through iterative retrieval and verification.
Search-o1 \citep{li-etal-2025-search} extends this paradigm by interleaving agentic search with reasoning while maintaining a condensed summary of retrieved documents as its knowledge state.
Beyond such textual representations, several methods introduce structured graph-based representations to manage growing context.
SG-Prompt \citep{li2023leveraging} and ERA-CoT \citep{liu-etal-2024-era} adopt structured prompting strategies for multi-step reasoning.
KnowTrace \citep{li2025knowtrace} summarizes each step's knowledge into a structured graph, keeping reasoning state compact.
RAS \citep{jiang2025ras} maintains a graph-structured knowledge state with iterative planning, while MultiCube-RAG \citep{shi2026multicuberag} employs hierarchical multi-dimensional indexing for iterative subquery retrieval.
Despite these advances, step-wise methods face two limitations: they lack global foresight, deciding the next subquery based only on the current state without a coherent plan, and they rely on the LLM to determine when to stop, so intermediate errors can propagate undetected.

\subsection{Plan-based Approaches}
Plan-based methods construct a global reasoning blueprint before execution, fixing the number of steps and thus determining when to stop at planning time.
PlanRAG \citep{lee2024planrag} and Plan*RAG \citep{verma2024planxrag} generate linear or DAG-structured plans from parametric knowledge without external context, while ReWOO \citep{xu2023rewoo} and LPKG \citep{wang2024lpkg} enhance reliability via in-context learning and task-specific training.
PAR-RAG \citep{zhang2025parrag}, ComposeRAG \citep{chen2025composerag}, and MA-RAG \citep{nguyen2025marag} adopt modular architectures with conditional or selective execution. Grounding planning in retrieval has been explored by STORM \citep{shao2024storm} and \citet{godbole2024planbased} for article generation. However, these grounded methods use preliminary retrieval to enrich the plan rather than to prune it, producing a complete reasoning trajectory regardless of whether the information is already covered. While grounding in retrieval is well-explored in step-wise methods, it remains underexplored in plan-based frameworks. GDP-RAG introduces gap-aware query decomposition: it identifies what is already known from preliminary retrieval and generates subqueries only for the information delta ($\Delta$), yielding a minimal skeletal trajectory rather than a complete reasoning chain.
In contrast to these methods, including \citet{godbole2024planbased} which grounds planning in retrieval yet still produces a complete reasoning path, GDP-RAG generates subqueries only for the missing information, so the plan itself is already minimal before execution begins.

\subsection{Retrieval and Verification}
Beyond query decomposition, multi-hop reasoning depends on accurate retrieval and answer verification at each step.
Standard RAG \citep{lewis2020rag} suffers from query ambiguity, motivating query expansion \citep{gao2023hyde,xiao2024knowledgeaware} and structured retrieval via knowledge graphs \citep{edge2024graphrag,gutierrez2024hipporag,li2026codrag}.
Robustness is further improved through filtering and verification \citep{yan2024corrective,asai2024selfrag,dhuliawala2023chain}.
GDP-RAG builds on the Act-Review-Update mechanism from PAR-RAG \citep{zhang2025parrag}. Act retrieves documents and generates a provisional answer, Review cross-verifies it via secondary retrieval, and Update commits the verified result to a persistent Trajectory Memory and refines the next sub-question based on accumulated evidence, providing runtime adaptability so that execution is not locked to the initial plan.

\subsection{Improving the Planner and Reasoner}
Methods for improving the planner and reasoner in multi-step RAG fall into training-free and training-based approaches.
Training-free approaches include prompting-based methods such as ReAct \citep{yao2023react} and IRCoT \citep{trivedi2023ircot}, and few-shot learning with planning demonstrations as in ReWOO \citep{xu2023rewoo}.
Training-based methods instead fine-tune the LLM, either via supervised learning, including LPKG \citep{wang2024lpkg} and Self-RAG \citep{asai2024selfrag}, or via reinforcement learning.
Search-R1 \citep{jin2025searchr1} introduces an RL framework that optimizes LLM reasoning with multi-turn search interactions through outcome-based rewards, R1-Searcher \citep{song2025r1searcher} proposes a two-stage outcome-based RL approach, and FrugalRAG \citep{java2025frugalrag} explores efficiency-oriented RL fine-tuning.
GDP-RAG is training-free, leveraging preliminary retrieval and gap-aware instructions to guide the planner toward a concise plan.

\section{Method}
\subsection{Problem Formulation}
\label{sec:problem_formulation}
We study efficient multi-step RAG.
Given a query $Q$ and a document collection $\mathcal{D}$, the goal is to produce an answer $A$ by constructing a reasoning trajectory $\mathcal{T}$ that integrates information from retrieved documents across multiple steps, such that:
\begin{equation}
A = \text{Answer}(Q, \mathcal{T}),
\end{equation}
where Trajectory $\mathcal{T} = \langle (q_1, d_1, a_1), \dots, (q_N, d_N, a_N) \rangle$ is defined as a sequence of $N$ steps. The plan-based method first decomposes $Q$ into a set of subqueries $\{q_1, \dots, q_N\}$. Subsequently, at each step $t \in \{1, \dots, N\}$, the system utilizes $q_t$ to retrieve a document set $d_t \subset \mathcal{D}$ and generates an intermediate answer $a_t$. The objective is to maximize the final answer $A$ quality, while minimizing computational cost, captured by minimizing the trajectory length $N$.
This requires a minimal yet sufficient set of subqueries that uncover the information required to answer $Q$, while avoiding redundant retrieval and reasoning.

\subsection{Framework of GDP-RAG}
\label{sec:GDP-RAG-framework}

\begin{figure}[t]
    \centering
    \includegraphics[width=\textwidth]{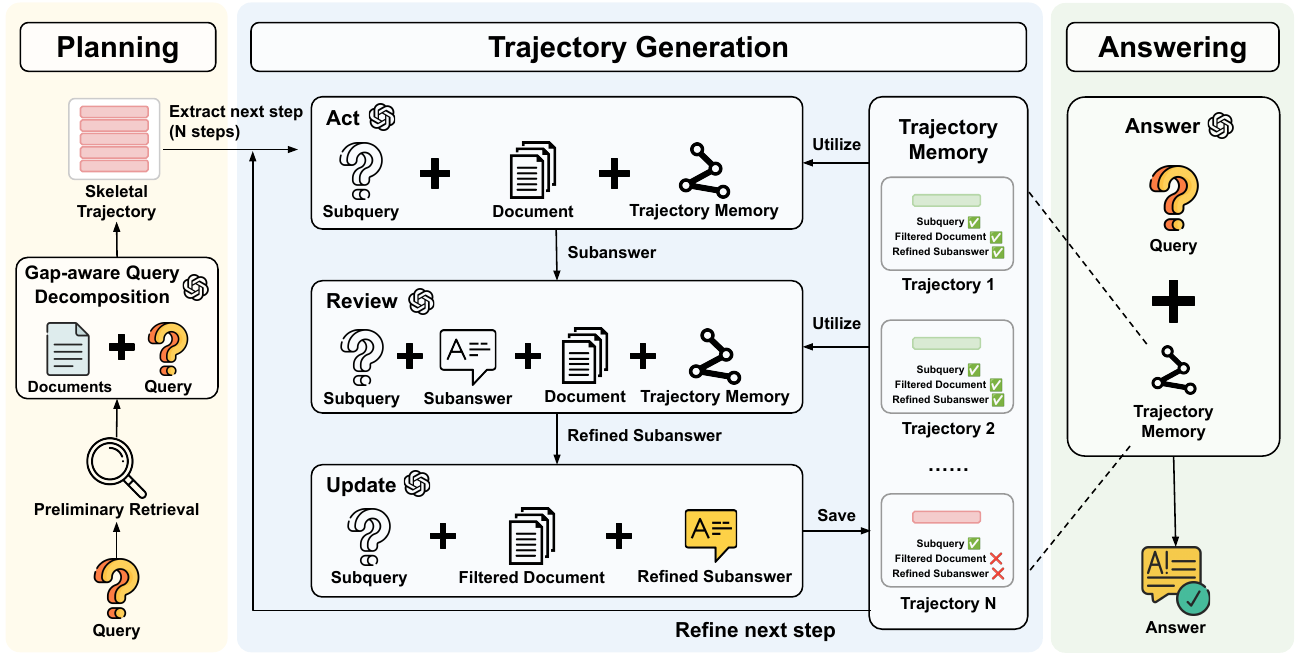}
    \caption{GDP-RAG framework. The workflow proceeds in three phases: (1) \textbf{Planning Phase}: Utilizes Grounded Delta Planning to generate a concise skeletal trajectory $\mathcal{T}_{skel}$, leveraging context from preliminary retrieval to prune redundant steps. (2) \textbf{Trajectory Generation Phase}: Executes an iterative Act-Review-Update loop, where each step interacts with Trajectory Memory and undergoes unconditional verification to transform the skeleton into a fully verified trajectory. (3) \textbf{Answering Phase}: Synthesizes the final response based on this comprehensive trajectory $\mathcal{T}$.}
\label{fig:workflow}
\end{figure}

As illustrated in Figure~\ref{fig:workflow}, GDP-RAG consists of three stages: Planning (§~\ref{subsubsec:grounded_planning}), Trajectory Generation (§~\ref{subsubsec:trajectory_gen}), and Answering (§~\ref{subsubsec:answer_derivation}).
Planning performs preliminary retrieval to obtain $D_{rel}$ and outputs a skeletal trajectory $\mathcal{T}_{skel}$ containing only essential subqueries.
Trajectory Generation executes and verifies each subquery in $\mathcal{T}_{skel}$, yielding a fully grounded trajectory $\mathcal{T}$.
Answering synthesizes the final answer $A$ from $\mathcal{T}$.

\subsubsection{Planning Phase}
\label{subsubsec:grounded_planning}

The Planning Phase decomposes $Q$ into a minimal set of subqueries that target only the information delta not yet covered by $D_{rel}$. This is realised through three design choices.

\paragraph{Preliminary Retrieval.}
Without retrieved context, the planner must rely solely on parametric knowledge, risking redundant or misguided subqueries.
To address this, the system first retrieves query-related context $D_{rel} \subset \mathcal{D}$ via coarse retrieval on $Q$.
$D_{rel}$ grounds the planner in concrete evidence before decomposition and defines what information is already available so that planning targets only the remaining gaps. We ablate passage quality in Section~\ref{sec:ablation} and show that relevant passages improve both step efficiency and accuracy.

\paragraph{Gap-aware Query Decomposition.}
Even with retrieved documents, the planner may still decompose the full query without recognizing what is already answered. To focus on the delta, the planning prompt enforces two instructions: (1) identify information available in $D_{rel}$, and (2) only create steps for information not in $D_{rel}$. This ensures the planner generates sub-questions only for $\Delta$. We ablate this in Section~\ref{sec:ablation} and show it primarily drives efficiency. The Direct Planning baseline omits $D_{rel}$ and the gap instructions. Full prompts for Direct and GDP-RAG are in Appendix Figures~\ref{fig:prompt_direct} and \ref{fig:prompt_grounded}.

\paragraph{Skeletal Trajectory.}
To organize what is already known and what still needs to be retrieved into a unified structure, the planner outputs a skeletal trajectory:
\begin{equation}
\mathcal{T}_{skel} = \mathrm{Plan}\big(Q, D_{rel}\big) = \langle (\theta_1, q_1), \dots, (\theta_n, q_n) \rangle,
\end{equation}
where each \texttt{Thought} $\theta_i$ preserves what evidence from $D_{rel}$ has already been obtained, and each subquery $q_i$ targets the information that is still missing.
This extends the standard trajectory tuple from $(q_t, d_t, a_t)$ to $(\theta_t, q_t, d_t, a_t)$.
Preserving $\theta_t$ serves two purposes: the planner must justify every subquery, which discourages unnecessary steps, and the evidence in $\theta_t$ is carried through Trajectory Memory to the final answering stage, so it is not lost during execution. PAR-RAG also generates Thoughts but discards them afterward. We ablate the Thought component in Section~\ref{sec:ablation} and show it is the primary driver of accuracy.

\paragraph{Advantages of Grounded Delta Planning. }
Together, these three choices yield three structural benefits (empirically validated in Section~\ref{sec:effectiveness}):
(1) \emph{Reasoning-step Compression}: $D_{rel}$ often already contains facts that would otherwise require dedicated steps. Since $\theta_t$ preserves this evidence through to the answering stage, those steps can be safely skipped without information loss.
(2) \emph{Over-planning Pruning}: Without retrieved evidence, planners rely on parametric knowledge and tend to hallucinate unnecessary steps. Grounding on $D_{rel}$ lets the planner verify assumptions against actual documents, eliminating subqueries with no factual basis.
(3) \emph{Under-planning Refinement}: Direct Planning decomposes the query into narrow, targeted subqueries that may miss relevant documents. Because $D_{rel}$ is retrieved using the original query, its broader semantics can retrieve entities and relations that targeted subqueries would miss, helping the planner capture dependencies that they would miss.

\subsubsection{Trajectory Generation Phase}
\label{subsubsec:trajectory_gen}
The skeletal plan $\mathcal{T}_{skel} = \langle (\theta_1, q_1), \dots, (\theta_N, q_N) \rangle$ defines the planned sequence of subqueries before execution. During execution, the system progressively grounds each subquery through an Act, Review, Update cycle adopted from PAR-RAG \citep{zhang2025parrag}, storing results in Trajectory Memory $\mathcal{M}_t = \langle (\theta_1, q_1, d_1, a_1), \dots, (\theta_t, q_t, d_t, a_t) \rangle$, which records the grounded execution state up to step $t$. In PAR-RAG, the plan module decides whether each subquery goes through the full Act, Review, Update cycle or is answered directly from Trajectory Memory. GDP-RAG instead executes every step through the full cycle, since Grounded Delta Planning already ensures that only genuine information gaps remain in $\mathcal{T}_{skel}$ (Appendix~\ref{sec:parrag_comparison}). Furthermore, Trajectory Memory $\mathcal{M}_{t-1}$ is provided to all three modules at each step, ensuring consistency across the reasoning process.

Each subquery targets a specific information gap that requires dedicated retrieval to resolve.
At each step $t$, the \textbf{Act} module retrieves relevant documents from $\mathcal{D}$ for the planned subquery $q_t$ and produces a provisional answer $\hat{a}_t$ conditioned on the query, the retrieved documents, and the current state of Trajectory Memory.
Formally:
\begin{equation}
\hat{a}_t = \mathrm{Act}\big(q_t, \mathcal{D}, \mathcal{M}_{t-1}\big),
\end{equation}
where $\mathcal{M}_{t-1}$ denotes the Trajectory Memory up to step $t-1$.
For factual transparency, we adopt a citation-enhanced generation strategy \citep{li2024citation}, in which $\hat{a}_t$ is generated by explicitly citing supporting documents from the retrieved set. At this stage, $\hat{a}_t$ remains provisional and has not yet been committed to the final reasoning trajectory.

However, a single retrieval pass may miss relevant documents or introduce hallucinations that propagate through subsequent steps.
To mitigate this, the \textbf{Review} module validates $\hat{a}_t$ before it is finalized. The system constructs a verification query by concatenating $q_t$ and $\hat{a}_t$, and performs a secondary retrieval over $\mathcal{D}$ to obtain targeted documents. The provisional answer is then examined against the newly retrieved documents to identify and correct factual inconsistencies.
Formally, the Review operation is defined as:
\begin{equation}
a_t = \mathrm{Review}\big(q_t, \hat{a}_t, \mathcal{D}, \mathcal{M}_{t-1}\big),
\end{equation}
where $a_t$ denotes the validated intermediate answer.

Because static plans cannot adapt to information discovered during execution, later subqueries may be redundant or poorly targeted.
The \textbf{Update} module addresses this by committing $(\theta_t, q_t, d_t, a_t)$ to Trajectory Memory, where $d_t$ is documents cited during Act and Review, and refines the next subquery for runtime adaptability:
\begin{equation}
(\mathcal{M}_t, q_{t+1}) = \mathrm{Update}(q_t, a_t, d_t, \mathcal{M}_{t-1})
\end{equation} where $\mathcal{M}_t$ represents the updated Trajectory Memory and $q_{t+1}$ is the refined subquery for the next step. 
Using the newly updated trajectory memory, the system refines $q_{t+1}$ to incorporate the latest findings before advancing to the subsequent cycle.
The cycle repeats until all $N$ steps are completed, yielding the fully grounded trajectory:
\begin{equation}
\mathcal{T} = \mathcal{M}_N = \langle (\theta_1, q_1, d_1, a_1), \dots, (\theta_N, q_N, d_N, a_N) \rangle
\end{equation}
This verified trajectory, containing both the reasoning structure and supporting citations, then serves as the comprehensive context for the final stage.

\subsubsection{Answering Phase}
\label{subsubsec:answer_derivation}
The system synthesizes the completed trajectory $\mathcal{T}$ with query $Q$ to generate the final answer $A$ (Section~\ref{sec:problem_formulation}), leveraging the verified tuples $(\theta_t, q_t, d_t, a_t)$ in Trajectory Memory to produce a response that is logically consistent and strictly supported by retrieved documents.

\section{Experiment Setup}

\paragraph{Datasets.}
We evaluate on three multi-hop QA benchmarks, denoted l1--l3 in all tables: HotpotQA \citep{yang2018hotpotqa}, 2WikiMultiHopQA \citep{xanh2020_2wikimultihop}, and MuSiQue \citep{trivedi2022musique}.
We use GPT-4.1-mini, the same model used by all methods for generation, to filter out questions answerable by parametric knowledge alone, and sample a balanced set across complexities.
We additionally report results on the full unfiltered splits in Appendix~\ref{appendix:filtered_unfiltered}, where the method ranking is preserved, confirming the comparison is not an artifact of filtering.
The resulting evaluation sets contain 200 questions per 2/3/4-hop category for HotpotQA and MuSiQue, and a 300/0/300 (2-hop/3-hop/4-hop) distribution for 2WikiMultiHopQA.
For HotpotQA the hop label denotes the number of supporting documents, since HotpotQA's questions are inherently 2-hop.
Details in Appendix~\ref{apendix:dataset}.

\begin{table*}[h]
\centering
\small
\setlength{\tabcolsep}{3.5pt}
\resizebox{\textwidth}{!}{%
\begin{tabular}{l|cccc|cccc|cccc|cccc}
\toprule
\multirow{3}{*}{Method} & \multicolumn{4}{c|}{Efficiency} & \multicolumn{4}{c|}{Effectiveness} & \multicolumn{8}{c}{Cost} \\
\cmidrule(lr){2-5}\cmidrule(lr){6-9}\cmidrule(lr){10-17}
& \multicolumn{4}{c|}{cost-of-pass$\downarrow$} & \multicolumn{4}{c|}{Acc./\%$\uparrow$} & \multicolumn{4}{c|}{Cost (\textcent{})$\downarrow$} & \multicolumn{4}{c}{\#Tokens$\downarrow$} \\
\cmidrule(lr){2-5}\cmidrule(lr){6-9}\cmidrule(lr){10-13}\cmidrule(lr){14-17}
& all & l1 & l2 & l3 & all & l1 & l2 & l3 & all & l1 & l2 & l3 & all & l1 & l2 & l3 \\
\midrule
Standard RAG & \textbf{0.14} & \textbf{0.05} & \textbf{0.09} & \textbf{0.29} & 37.86 & 56.41 & 31.89 & 25.26 & \textbf{0.04} & \textbf{0.03} & \textbf{0.03} & \textbf{0.07} & \textbf{1077} & \textbf{696} & \textbf{726} & \textbf{1809} \\
IRCoT & \underline{0.17} & \underline{0.07} & \underline{0.11} & \underline{0.31} & 44.56 & 54.11 & 52.99 & 26.58 & \underline{0.06} & \underline{0.04} & \underline{0.06} & \underline{0.08} & \underline{1284} & \underline{810} & \underline{1204} & \underline{1839} \\
KnowTrace & 1.57 & 0.68 & 0.51 & 3.52 & 51.65 & 55.66 & 68.31 & 30.99 & 0.61 & 0.38 & 0.35 & 1.09 & 13519 & 8296 & 7692 & 24569 \\
Search-o1 & 0.49 & 0.30 & 0.34 & 0.81 & 50.73 & 51.03 & \textbf{72.66} & 28.51 & 0.21 & 0.16 & 0.25 & 0.23 & 4095 & 2881 & 5172 & 4233 \\
PAR-RAG & 0.65 & 0.39 & 0.43 & 1.14 & \underline{58.20} & \underline{62.79} & 69.10 & \underline{42.70} & 0.34 & 0.24 & 0.30 & 0.49 & 7112 & 4856 & 5956 & 10523 \\
Godbole et al. & 1.12 & 0.54 & 0.64 & 2.16 & 46.28 & 59.42 & 44.83 & 34.59 & 0.45 & 0.32 & 0.29 & 0.75 & 9606 & 6511 & 5964 & 16343 \\
\midrule
GDP-RAG & 0.51 & 0.25 & 0.30 & 0.98 & \textbf{60.63} & \textbf{64.75} & \underline{72.62} & \textbf{44.53} & 0.27 & 0.16 & 0.22 & 0.44 & 5717 & 3271 & 4460 & 9420 \\
\bottomrule
\end{tabular}
}
\caption{Main results. all = average of the three per-dataset values, applied uniformly to all metrics. l1/l2/l3 = HotpotQA/2WikiMultiHopQA/MusiQue, each averaged over hops. \textbf{Bold} = best, \underline{underline} = second best.}
\label{tab:main-results}
\end{table*}

\begin{figure}[h]
    \centering
    
    \begin{subfigure}[b]{0.48\textwidth}
        \centering
        \includegraphics[width=\linewidth]{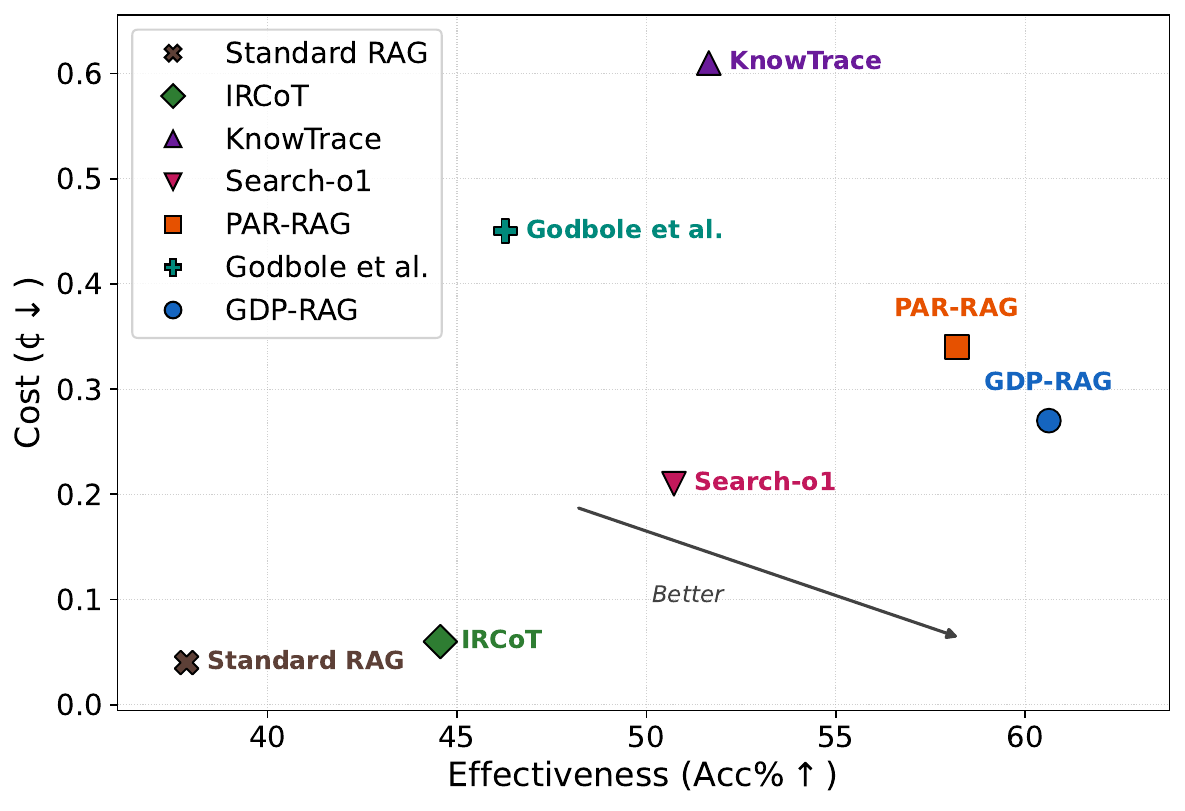}
        \caption{Baseline Comparison}
        \label{fig:eff_eff}
    \end{subfigure}
    \hfill 
    \begin{subfigure}[b]{0.48\textwidth}
        \centering
        \includegraphics[width=\linewidth]{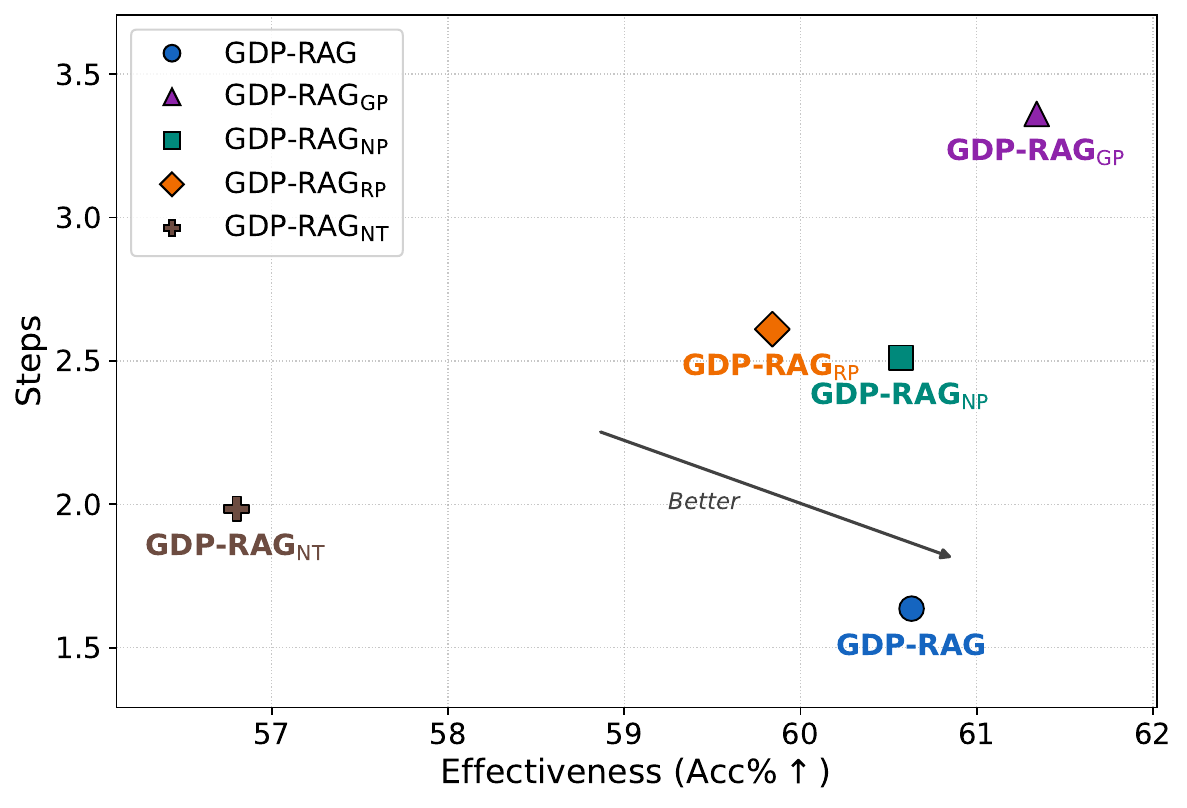}
        \caption{Planning Ablation}
        \label{fig:Plan_abla}
    \end{subfigure}
    \caption{Experimental results of the proposed GDPRAG framework. (a) shows the trade-off between effectiveness and efficiency, while (b) illustrates the impact of planning steps.}
    
    \label{fig:combined_results}
\end{figure}

\paragraph{Evaluation Metrics.}
Following the methodology of previous work \citet{wang2025efficientagents}, we evaluate the RAG system along three axes. 
For effectiveness, we report Accuracy (Acc), the average of Exact Match (EM), F1 score (F1), and Substring Match (SM).
EM checks whether the prediction exactly matches the ground truth, F1 measures token-level overlap, and SM checks whether the ground truth appears as a substring of the prediction. Per-metric definitions are in Appendix~\ref{appendix:effectiveness}, and per-dataset EM/F1/SM scores in Table~\ref{tab:appendix-effectiveness-by-dataset}. For cost, we report Cost (\textcent{}) and \#Tokens, the monetary cost and total tokens consumed per query. For efficiency, we report Cost-of-pass (\textcent{}), the expected cost to produce one correct answer, jointly capturing cost and accuracy in a single figure and serving as our primary metric.
Detailed formulas and pricing are provided in Appendix~\ref{appendix:efficiency}.

\paragraph{Baselines.}
We evaluate against five baselines in two categories.
\textit{Step-wise methods} interleave retrieval and reasoning one step at a time:
IRCoT \citep{trivedi2023ircot} with chain-of-thought reasoning,
KnowTrace \citep{li2025knowtrace} with an iteratively built knowledge graph,
and Search-o1 \citep{li-etal-2025-search} with a condensed summary of retrieved documents as its knowledge state.
\textit{Plan-based methods} build a plan before execution:
PAR-RAG \citep{zhang2025parrag} generates all sub-questions upfront from parametric knowledge,
and Godbole et al.\ \citep{godbole2024planbased} derive queries from an outline grounded in initial retrieval.
We reproduce KnowTrace in its inference-time KG-tracing form only, without the self-bootstrapping fine-tuning, to keep all comparisons training-free.
As \citet{godbole2024planbased} was designed for article writing, we adapt it to QA by keeping only its first planning stage and pairing it with our execution and answering modules, which isolates the planning strategy: it constructs full reasoning paths, whereas GDP-RAG plans only for the information delta.

\paragraph{Implementation Details.}
All methods use GPT-4.1-mini \citep{openai_gpt41mini_2024} for planning and generation. Documents are encoded with BAAI/bge-m3 \citep{chen2024bge}; top-30 candidates are reranked by BAAI/reranker-large-v2-m3 \citep{baai2024rerankerm3} to yield the final top-10. Step-wise methods (IRCoT, KnowTrace, Search-o1) are capped at 10 reasoning steps. All experiments run on a single NVIDIA RTX 3090 GPU.

\section{Main Results}
Table~\ref{tab:main-results} and Figure~\ref{fig:eff_eff} present the main results. GDP-RAG attains the highest overall accuracy (60.63\%), surpassing the strongest baseline PAR-RAG by 2.43 percentage points. It attains the best accuracy on HotpotQA (64.75\%) and MusiQue (44.53\%), and ranks a close second on 2Wiki (72.62\%, just behind Search-o1's 72.66\%). The gains confirm that gap-conditioned planning not only reduces unnecessary retrieval but also improves answer quality by focusing on genuinely missing information. A paired significance test (paired $t$-test and a 10{,}000-sample bootstrap) confirms that GDP-RAG's improvement over the strongest baseline PAR-RAG is statistically significant on EM and F1 ($p<0.001$); the full per-baseline analysis is reported in Appendix~\ref{appendix:significance}.

On efficiency, GDP-RAG lies on the Pareto frontier: no method reaches both higher accuracy and lower cost, while the only cheaper methods, Standard RAG (0.04\textcent{}), IRCoT (0.06\textcent{}), and Search-o1 (0.21\textcent{}), trail it by 10 to 23 accuracy points. Cost-of-pass, our primary metric, summarizes this: GDP-RAG scores 0.51, and the only lower scores (Standard RAG 0.14, IRCoT 0.17, Search-o1 0.49) belong to these low-accuracy methods, placing GDP-RAG in the high-accuracy, low-cost corner of Figure~\ref{fig:eff_eff}. Finally, we examine robustness to model scale: under the weaker GPT-4.1-nano, every method loses accuracy and GDP-RAG's lead narrows, yet it stays cheaper than the strongest baseline PAR-RAG (0.07 vs.\ 0.09\textcent{}) at comparable accuracy (40.00\% vs.\ 41.38\%; Appendix~\ref{appendix:model_scale}).

\begin{table*}[t]
\centering
\small
\setlength{\tabcolsep}{3.5pt}
\resizebox{\textwidth}{!}{%
\begin{tabular}{l|cccc|cccc|cccc|cccc}
\toprule
\multirow{3}{*}{Method} & \multicolumn{4}{c|}{Efficiency} & \multicolumn{4}{c|}{Effectiveness} & \multicolumn{8}{c}{Cost} \\
\cmidrule(lr){2-5}\cmidrule(lr){6-9}\cmidrule(lr){10-17}
& \multicolumn{4}{c|}{cost-of-pass$\downarrow$} & \multicolumn{4}{c|}{Acc./\%$\uparrow$} & \multicolumn{4}{c|}{Cost (\textcent{})$\downarrow$} & \multicolumn{4}{c}{\#Tokens$\downarrow$} \\
\cmidrule(lr){2-5}\cmidrule(lr){6-9}\cmidrule(lr){10-13}\cmidrule(lr){14-17}
& all & l1 & l2 & l3 & all & l1 & l2 & l3 & all & l1 & l2 & l3 & all & l1 & l2 & l3 \\
\midrule
\multicolumn{17}{l}{\textbf{Planning Ablation}} \\
\midrule
GDP-RAG & \textbf{0.51} & \textbf{0.25} & \textbf{0.30} & \textbf{0.98} & \underline{60.63} & 64.75 & \underline{72.62} & 44.53 & \textbf{0.27} & \textbf{0.16} & \underline{0.22} & \textbf{0.44} & \textbf{5717} & \textbf{3271} & \textbf{4460} & \textbf{9420} \\
GDP-RAG$_{GP}$ & 0.93 & 0.57 & 0.51 & 1.72 & \textbf{61.34} & \textbf{65.63} & \textbf{75.31} & 43.07 & 0.50 & 0.37 & 0.38 & 0.74 & 10512 & 7504 & 7862 & 16169 \\
GDP-RAG$_{NP}$ & \underline{0.59} & 0.33 & 0.44 & \underline{1.00} & 60.57 & \underline{64.90} & 71.20 & \textbf{45.61} & 0.33 & 0.21 & 0.32 & \underline{0.46} & 6857 & 4288 & 6442 & \underline{9841} \\
GDP-RAG$_{RP}$ & 0.69 & 0.39 & 0.48 & 1.21 & 59.84 & 63.37 & 71.21 & \underline{44.95} & 0.38 & 0.25 & 0.34 & 0.54 & 7962 & 5013 & 7008 & 11866 \\
GDP-RAG$_{NT}$ & 0.60 & \underline{0.27} & \underline{0.31} & 1.23 & 56.80 & 62.40 & 67.19 & 40.82 & \underline{0.29} & \underline{0.17} & \textbf{0.21} & 0.50 & \underline{6491} & \underline{3593} & \underline{4518} & 11362 \\
\midrule
\multicolumn{17}{l}{\textbf{Trajectory Generation Ablation}} \\
\midrule
GDP-RAG & 0.51 & 0.25 & 0.30 & 0.98 & \textbf{60.63} & \textbf{64.75} & \underline{72.62} & \textbf{44.53} & 0.27 & 0.16 & 0.22 & 0.44 & 5717 & 3271 & 4460 & 9420 \\
GDP-RAG$_{NR}$ & \underline{0.41} & \underline{0.21} & \underline{0.24} & \textbf{0.78} & 59.41 & 62.99 & 71.50 & \underline{43.75} & \underline{0.21} & \underline{0.13} & \underline{0.17} & \underline{0.34} & \underline{4512} & \underline{2669} & \underline{3426} & \underline{7440} \\
GDP-RAG$_{NU}$ & 0.51 & 0.25 & 0.27 & 1.01 & \underline{59.77} & \underline{64.08} & \textbf{73.89} & 41.33 & 0.26 & 0.16 & 0.20 & 0.42 & 5511 & 3251 & 4177 & 9106 \\
\bottomrule
\end{tabular}
}
\caption{Ablation study. Planning ablation (top) and Trajectory Generation ablation (bottom).}
\label{tab:ablation}
\end{table*}

\section{Ablation Study}
\label{sec:ablation}
To analyze each component in GDP-RAG, we conduct an ablation study of both the Planning and Trajectory Generation phases. In the Planning phase, we ablate the three design choices introduced in §~\ref{subsubsec:grounded_planning}: the quality of preliminary $D_{rel}$, the gap-conditioning instruction, and the skeletal trajectory format. In the Trajectory Generation phase, we ablate the two-stage retrieval (Review) mechanisms and sub-question refinement (Update).

\subsection{Planning Phase}
Since all planning variants share the same trajectory generation phase, differences reflect purely planning-phase effects (Table~\ref{tab:ablation}). Figure~\ref{fig:Plan_abla} visualizes the step count vs.\ accuracy trade-off, where GDP-RAG occupies the bottom-right (fewest steps, high accuracy).

\paragraph{Preliminary Retrieval.} 
We compare GDP-RAG against two variants: GDP-RAG$_{NP}$ (No Passages), which removes all passages from the planner input, and GDP-RAG$_{RP}$ (Random Passages), which replaces retrieved passages with 10 randomly sampled documents.
Without passages (NP), accuracy is nearly unchanged (60.57\%) but cost rises 22\% (0.33 vs.\ 0.27~\textcent{}/query) and steps increase to 2.51. With random passages (RP), accuracy drops further (59.84\%) and cost is even higher (0.38~\textcent{}/query, 2.61 steps). Notably, RP is worse than NP on both axes, showing that irrelevant passages are worse than no passages at all.

\paragraph{Gap-aware Query Decomposition.}
We compare GDP-RAG against GDP-RAG$_{GP}$ (Grounded Planning), which removes the gap-conditioning instruction. The planner still receives passages and generates \texttt{Thought}s, but without gap instructions.
Without gap-conditioning instruction, accuracy (+0.71\%) is marginally higher but cost rises 85\% (0.50 vs.\ 0.27~\textcent{}), generating 3.36 steps versus 1.64 due to redundant sub-questions. Even GDP-RAG$_{NP}$ generates fewer steps (2.51) than GP (3.36), confirming that gap instruction drives step reduction independent of passages.

\paragraph{Skeletal Trajectory.} 
We compare GDP-RAG against GDP-RAG$_{NT}$ (No Thought), which removes the Thought component ($\theta_t$) from the skeletal trajectory, reducing each entry to only $q_t$. Without $\theta_t$, evidence from $D_{rel}$ is not preserved through execution. NT causes the largest accuracy drop among all planning ablations (56.80\% vs.\ 60.63\%), while cost remains similar (0.29 vs.\ 0.27~\textcent{}). The skeletal trajectory format is critical for accuracy: it improves sub-question quality rather than merely reducing step count.

\subsection{Trajectory Generation Phase}
To evaluate the trajectory generation phase, we ablate the Review and Update modules while keeping the planning stage constant (Table~\ref{tab:ablation}). This setup isolates the effects of each module on reasoning quality and computational cost.

\paragraph{Review.} 
We compare GDP-RAG against GDP-RAG$_{NR}$ (No Review), which uses the provisional answer directly without secondary retrieval or verification. While NR reduces cost-of-pass by 20\% (0.41 vs.\ 0.51 \textcent{}) and tokens by 21\%, it causes only a marginal accuracy loss (59.41\% vs.\ 60.63\%). This indicates that Review is the dominant cost component in trajectory generation but contributes modestly to final accuracy.

\paragraph{Update.} 
We compare GDP-RAG against GDP-RAG$_{NU}$ (No Update), which disables the refinement of planned sub-questions based on previous trajectory steps. NU yields negligible cost savings (around 4\%) but results in a consistent accuracy drop (59.77\% vs.\ 60.63\%). Since sub-question refinement requires only one additional LLM call per step, Update is essentially a low-cost yet effective mechanism for maintaining runtime adaptability.



Update should always be included as it requires only one additional LLM call per step while providing runtime adaptability. Review presents a cost-accuracy trade-off: it dominates cost but contributes modestly to accuracy, exposing two operating points --- full GDP-RAG for higher accuracy and GDP-RAG$_{NR}$ for lower cost.

\begin{figure}[t]
\centering
\begin{subfigure}[b]{0.48\linewidth}
    \centering
    \includegraphics[width=\linewidth]{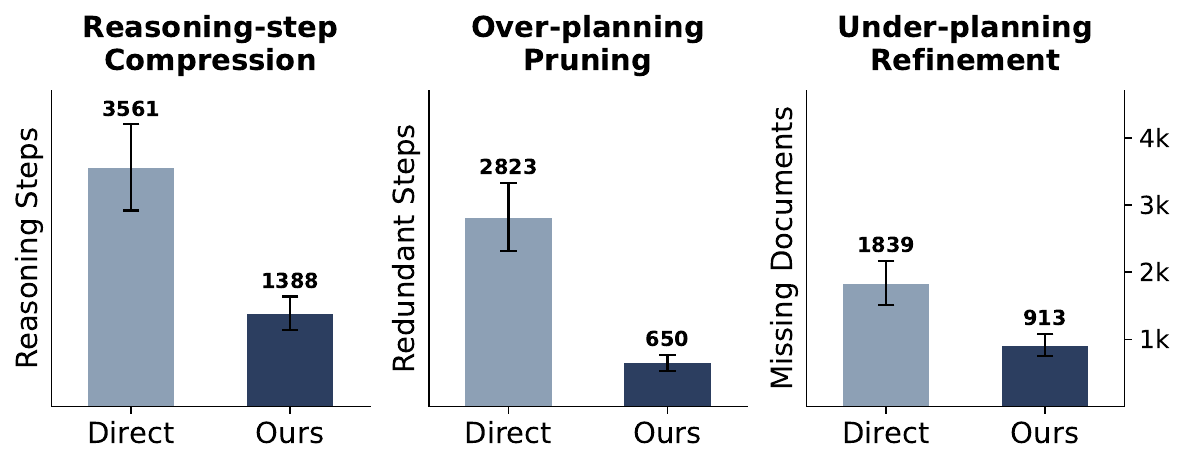}
    \caption{Reasoning Steps, Redundant Steps, and Missing Documents for Direct vs. Ours, evaluated 
  by LLM-based alignment with 82\% human-verified accuracy (details in 
  Appendix~\ref{sec:grounded_planning_analysis}).}
    \label{fig:planning_stats}
\end{subfigure}
\hfill 
\begin{subfigure}[b]{0.48\linewidth}
    \centering
    \includegraphics[width=\linewidth]{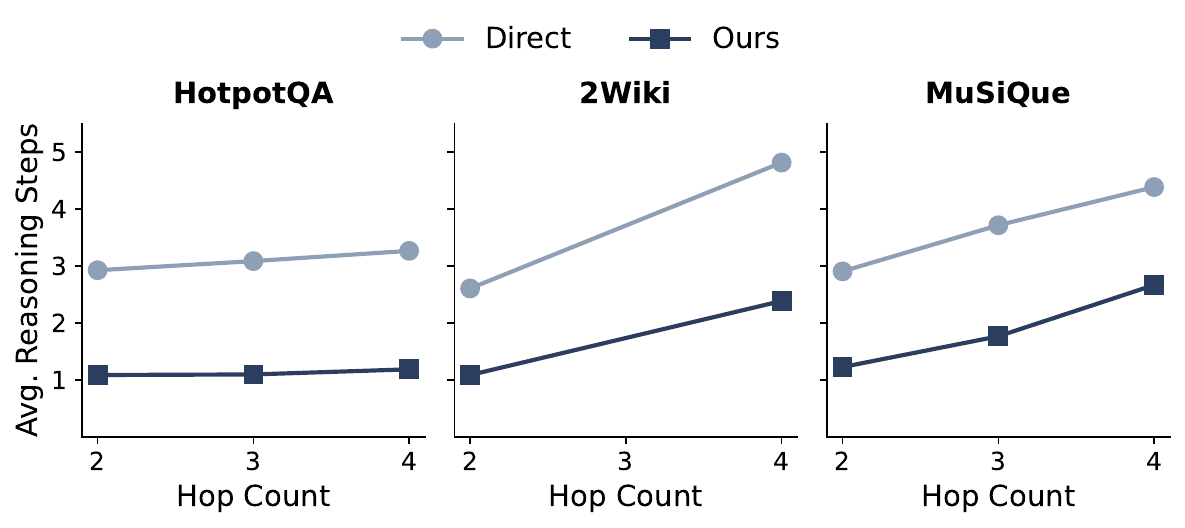}
    \caption{Average reasoning steps across 2- to 4-hop queries for Direct Planning vs. Grounded Delta Planning on three datasets. The gap widens as hop count increases.}
    \label{fig:hop}
\end{subfigure}

\caption{Comparison of planning statistics (a) and reasoning complexity by query hops (b).}
\label{fig:merged_planning}
\end{figure}



\section{Effectiveness Analysis of GDP-RAG}
\label{sec:effectiveness}
Beyond aggregate accuracy, we examine how Grounded Delta Planning reshapes the planning process itself. We compare against Direct Planning (without preliminary retrieval and gap-aware instruction). A qualitative case study is provided in Appendix~\ref{appendix:case-study}, Table~\ref{table:case-study}.

\subsection{Compression, Pruning, and Refinement.}
Figure~\ref{fig:planning_stats} quantifies the impact: an LLM-based evaluator labels each planned step as a Reasoning Step that targets a gold document, a Redundant Step that over-plans, or a Missing Document not covered by the plan, with 82\% agreement against human review on a 10\% sample. Classification criteria are detailed in Appendix~\ref{sec:grounded_planning_analysis}.

GDP-RAG improves on all three dimensions: Reasoning-step Compression cuts valid steps from 3{,}561 to 1{,}388 (61.0\%), Over-planning Pruning cuts redundant steps from 2{,}823 to 650 (77.0\%), and Under-planning Refinement lowers missing gold documents from 1{,}839 to 913 (50.4\%), confirming that grounding the planner both reduces overhead and strengthens coverage. To prove the gains are not an artifact of a weak planner, we use GPT-5.5 as the LLM in Appendix~\ref{appendix:plan_scale} and preserve the compression and pruning improvements.

The compression effect is governed by how much gold evidence the preliminary retrieval already covers. Its recall@10 declines as the hop count grows, averaging 0.61 across datasets (per-dataset, per-hop breakdown in Appendix Table~\ref{tab:recall-hop}). Because each reasoning step typically resolves a single passage, the fraction of steps the planner can compress tracks this coverage: the 61.0\% Reasoning-step Compression closely matches the 0.61 average recall.

\subsection{Efficiency under Increasing Query Complexity.}
Figure~\ref{fig:hop} compares average reasoning steps across 2- to 4-hop queries. Grounded Delta Planning consistently requires fewer steps, with the gap widening as complexity increases.
On HotpotQA and 2Wiki, Grounded Delta Planning maintains nearly constant step count, amortizing additional hops.
On MuSiQue, although steps increase linearly, GDP-RAG maintains an efficiency advantage.
These results show that Grounded Delta Planning adapts to task complexity, pruning redundant steps while retaining necessary reasoning.

\section{Conclusion}
We presented GDP-RAG, a plan-based multi-hop RAG framework that retrieves first and plans only for the information delta ($\Delta$) through preliminary retrieval, gap-aware query decomposition, and a skeletal trajectory. An Act-Review-Update cycle grounds each subquery with runtime adaptability. 
Experiments on HotpotQA, 2WikiMultiHopQA, and MuSiQue show that GDP-RAG achieves the highest accuracy among all compared systems with a cost-of-pass 22\% lower than PAR-RAG and 68\% lower than KnowTrace. Ablations confirm that each component contributes: preliminary retrieval improves step efficiency, the gap-aware query decomposition drives efficiency, and the skeletal trajectory drives accuracy.

\section{Limitations and Future Work}
Our evaluation targets short-form multi-hop QA, where information gaps are defined over factual entities and relations and answers are short and verifiable. Open-ended generation, long-form synthesis, and multi-modal retrieval would require a different notion of the information delta and a different evaluation metric. Extending GDP-RAG to such open-ended and agentic settings is a natural direction for future work.

\section*{Acknowledgments}

This work was supported by the National Science 
and Technology Council (NSTC), Taiwan, under Grant No. 
113-2740-H-002-001-MY3, ``TAIHUCAIS: TAIwan HUmanities 
Conversational AI Knowledge Discovery System.''
This work was also financially supported by the National Science and Technology
Council (NSTC) in Taiwan, under Grant 114-2628-E-002-022.
This work was supported by the Ministry of Education (MOE) of Taiwan under
the project Taiwan Centers of Excellence in Artificial Intelligence,
through the NTU Artificial Intelligence Center of Research Excellence (NTU AI-CoRE). 
We also thank the National Center for High-performance 
Computing (NCHC) of the National Applied Research 
Laboratories (NARLabs), Taiwan, for providing computational 
and storage resources.



\bibliography{colm2026_conference}
\bibliographystyle{colm2026_conference}

\appendix
\section{Appendix}

\subsection{Prompt Design}
This section details the transition from the baseline Direct Planning prompt to our proposed Grounded Planning formulation.

\begin{figure*}[h]
    \centering
    \includegraphics[width=\linewidth]{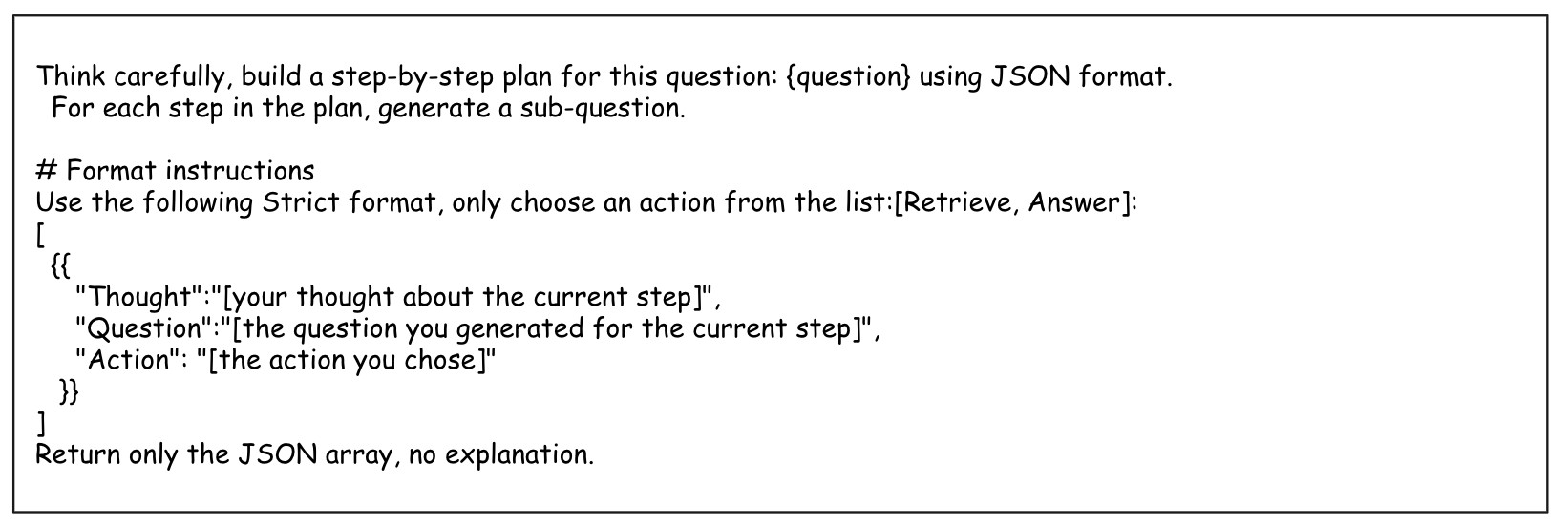}
    \caption{Prompt of Direct Planning.}
    \label{fig:prompt_direct}
\end{figure*}

\begin{figure*}[h]
    \centering
    \includegraphics[width=\linewidth]{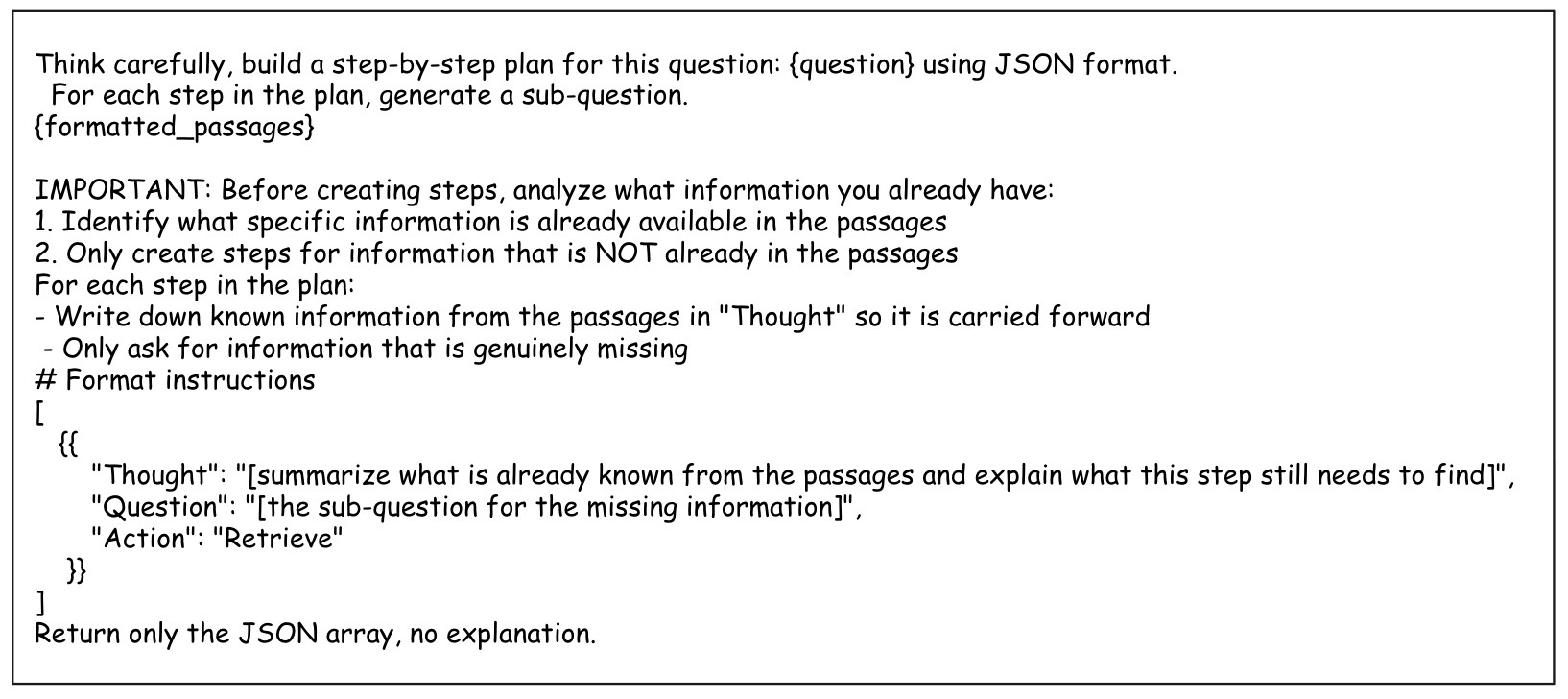}
    \caption{Prompt of Grounded Delta Planning.}
    \label{fig:prompt_grounded}
\end{figure*}

\paragraph{Direct Planning Prompt} 
\label{sec:direct_prompt}
As shown in Figure \ref{fig:prompt_direct}, our Direct Planning prompt is adapted from PAR-RAG, requiring the LLM to generate a structured JSON plan. For each reasoning step, the model must produce a \texttt{Thought} process and a specific subquery, while restricting its choice of actions to a predefined list: [\texttt{Retrieve}, \texttt{Answer}]. These generated actions serve as the foundational candidates for our subsequent conditional execution mechanism, allowing the system to determine which steps require external grounding. 

Retrieve: This action triggers the full Act-Review-Update cycle, where the system actively gathers new information and verifies its relevance to the reasoning trajectory.

Answer: This action indicates that the current trajectory memory contains sufficient information; the model then formulates a response based on it without further searching.

\paragraph{Grounded Delta Planning Prompt}
As shown in Figure~\ref{fig:prompt_grounded}, the Grounded Delta Planning prompt incorporates preliminary retrieval results and enforces two key instructions: (1) identify what information is already available in the retrieved passages, and (2) only create steps for information not already present. The resulting JSON output requires a Thought field to justify why specific information is not yet available and a Question field to define the missing details. Unlike the Direct Planning prompt, no predefined action list is provided.

\subsection{Comparison with PAR-RAG}
\label{sec:parrag_comparison}
Although GDP-RAG builds upon PAR-RAG \citep{zhang2025parrag}, the two frameworks differ in a key design choice: what to execute.
PAR-RAG generates a full reasoning trajectory from parametric knowledge and conditionally decides, at each step, whether to retrieve externally or answer from previously generated intermediate results (see Appendix~\ref{sec:direct_prompt}).
GDP-RAG inverts this logic: because Grounded Delta Planning already eliminates redundant and trivial steps during planning, every subquery in $\mathcal{T}_{skel}$ corresponds to a genuine information gap.
The system therefore passes through the full Act--Review--Update loop every planned step, increasing reliability by ensuring that no intermediate answer relies on unverified parametric reasoning.
GDP-RAG further strengthens consistency by sharing Trajectory Memory across the Act and Review modules and by committing all cited documents from both phases. Only documents that directly support each intermediate answer $a_t$ are stored as $d_t$, yielding a compact and fully verifiable trajectory $\mathcal{T}$.

\subsection{Filtered vs.\ Unfiltered Evaluation}
\label{appendix:filtered_unfiltered}
To verify that knowledge-intensive filtering does not drive our results, we additionally evaluate on the full unfiltered splits (600 questions per dataset). Table~\ref{tab:filtered-unfiltered} reports accuracy and cost on both settings, averaged over the three datasets. The method ranking is identical across the two splits: GDP-RAG attains the highest accuracy on both, and every method's relative ordering is preserved. We adopt the filtered set as the main setting because it removes questions answerable without retrieval, yielding cleaner ablations. The unfiltered split confirms the comparison is not an artifact of filtering.

\begin{table}[h]
\centering
\small
\setlength{\tabcolsep}{5pt}
\begin{tabular}{l|cc|cc}
\toprule
\multirow{2}{*}{Method} & \multicolumn{2}{c|}{Filtered} & \multicolumn{2}{c}{Unfiltered} \\
\cmidrule(lr){2-3}\cmidrule(lr){4-5}
& Acc\%$\uparrow$ & Cost\,\textcent{}$\downarrow$ & Acc\%$\uparrow$ & Cost\,\textcent{}$\downarrow$ \\
\midrule
Standard RAG   & 37.85 & 0.04 & 46.17 & 0.04 \\
IRCoT          & 44.56 & 0.06 & 50.25 & 0.06 \\
KnowTrace      & 51.65 & 0.61 & 52.47 & 0.57 \\
Godbole et al. & 46.28 & 0.45 & 52.16 & 0.45 \\
PAR-RAG        & 58.20 & 0.34 & 62.87 & 0.35 \\
GDP-RAG        & \textbf{60.63} & 0.27 & \textbf{64.09} & 0.27 \\
\bottomrule
\end{tabular}
\caption{Filtered vs.\ unfiltered evaluation. Each cell is accuracy (\%) and cost (\textcent{}/query), averaged over the three datasets and 600 questions each on the unfiltered split. The method ranking is identical across both settings.}
\label{tab:filtered-unfiltered}
\end{table}

\subsection{Dataset Curation Two Stage Processing} \label{apendix:dataset}
\paragraph{Knowledge-Intensive Filtering.} To ensure retrieval dependency, we exclude questions solvable by the LLM’s parametric knowledge (Table \ref{tab:relax_match_error_rate}). We identify such cases via Substring Match and evaluate only the remaining "Wrong" subset.

\begin{table}[h]
\centering
\small
\setlength{\tabcolsep}{4pt}
\renewcommand{\arraystretch}{1}
\begin{tabular}{l c c }
\toprule
\textbf{QA Dataset} & \textbf{Wrong(\%)} & \textbf{Correct(\%)} \\
\midrule
HotpotQA & 62.38 & 37.62 \\
2Wiki & 63.80 & 36.20 \\
MuSiQue & 88.95 & 11.05 \\
\bottomrule
\end{tabular}
\caption{Justification for Dataset Filtering.}
\label{tab:relax_match_error_rate}
\end{table}

\paragraph{Hop-Balanced Sampling.} We sample 600 questions per dataset. HotpotQA and MuSiQue use a uniform 200/200/200 split for 2/3/4 hops. 2Wiki uses 300/0/300 due to lack of 3-hop data. This prevents lower-hop dominance and enables scalability evaluation.

\subsection{Effectiveness Metrics} \label{appendix:effectiveness}

\begin{table}[h]
\centering
\small
\setlength{\tabcolsep}{3.5pt}
\begin{tabular}{l l c c c}
\toprule
\multicolumn{4}{l}{\textbf{Question:} Whose father is older, Amy or Bob? \quad \textbf{Ground Truth :} Bob} \\
\midrule
\textbf{Predictions} & \textbf{Tokenization} & \textbf{EM} & \textbf{F1} & \textbf{SM} \\
\midrule
\textbf{1)} Bob. & Bob & 100\% & 100\% & 100\% \\
\textbf{2)} Bob's father is older. & Bob/'s/father/is/older & 0\% & 33\% & 100\% \\
\textbf{3)} Bob's father is older than John's.& Bob/'s/father/is/older/than/John/'s & 0\% & 22\% & 100\% \\
\bottomrule
\end{tabular}
\caption{Example of Quality Metric Calculation for Question Answering.}
\label{tab:metric_example}
\end{table}

\paragraph{Exact Match (EM)} It tests the final synthesis quality. This metric tells us if the LLM successfully converted all the retrieved and reasoned facts into the final answer. This is a binary metric that scores 100\% for a perfect match and 0\% otherwise. As seen in Table \ref{tab:metric_example}, Prediction 2 and 3 both scores 0\% EM; only Prediction 1 scores 100\%. 

\paragraph{F1 score} It reflects partial correctness while simultaneously penalizing the model for extraneous content. It is calculated by determining the token-level Precision and Recall of the predicted answer against the ground truth. The final score is the harmonic mean of these two values, calculated using the function: 
$$
{F1} = 2 \cdot \frac{Precision \cdot Recall}{Precision + Recall}
$$
As shown in Table \ref{tab:metric_example}, the score drops as prediction length increases. For instance, while Recall remains perfect, the Precision drops from $1$ (Prediction 1) to $1/5$ (Prediction 2) to $1/8$ (Prediction 3), causing F1 to decline from $100\%$ to $22\%$.

\paragraph{Substring Match (SM)} It relaxes the need for perfect phrasing and rewards factual completeness, even if the prediction includes additional context or words. This is a forgiving, binary metric that scores 100\% if the entire ground-truth answer is a substring within the predicted answer, and $0\%$ otherwise. As shown in Table \ref{tab:metric_example}, all three predictions contain the ground truth ("Bob"), so they all achieve an SM score of $\mathbf{100\%}$.

\begin{table*}[h]
\centering
\small
\setlength{\tabcolsep}{5pt}
\resizebox{\textwidth}{!}{%
\begin{tabular}{l|ccc|ccc|ccc}
\toprule
\multirow{2}{*}{Method} &
\multicolumn{3}{c|}{HotpotQA} &
\multicolumn{3}{c|}{2Wiki} &
\multicolumn{3}{c}{MusiQue} \\
\cmidrule(lr){2-4}\cmidrule(lr){5-7}\cmidrule(lr){8-10}
& EM & F1 & SM & EM & F1 & SM & EM & F1 & SM \\
\midrule
IRCoT      & 41.67 & 58.83 & 61.83 & 44.00 & 51.81 & 63.16 & 16.33 & 29.24 & 34.17 \\
KnowTrace  & 41.50 & 59.47 & 66.00 & 55.16 & \underline{69.93} & \textbf{79.83} & 23.33 & 34.65 & 35.00 \\
PARRAG     & \underline{47.67} & \underline{68.03} & \textbf{72.67} & \underline{60.16} & 69.48 & 77.66 & \underline{30.17} & \underline{46.93} & \textbf{51.00} \\
Godbole et al.    & 47.00 & 64.58 & 66.67 & 33.00 & 40.32 & 61.16 & 25.17 & 39.44 & 39.17 \\
GDPRAG     & \textbf{52.33} & \textbf{70.40} & \underline{71.50} & \textbf{66.67} & \textbf{73.53} & \underline{77.67} & \textbf{34.67} & \textbf{48.43} & \underline{50.50} \\
\bottomrule
\end{tabular}
}
\caption{Effectiveness breakdown by dataset. We report EM, F1, and SM (all in \%) for each method on HotpotQA, 2Wiki, and MusiQue.}
\label{tab:appendix-effectiveness-by-dataset}
\end{table*}

\subsection{Cost and Efficiency Metrics}
\label{appendix:efficiency}

\paragraph{Cost (\textcent).} The monetary cost of a single inference attempt, computed as
$C = n_{{in}} \cdot c_{{in}} + n_{{out}} \cdot c_{{out}}$,
where $n_{{in}}$ and $n_{{out}}$ are the input and output token counts, and $c_{{in}}$, $c_{{out}}$ are the per-token prices (GPT-4.1-mini: \$0.40/1M input, \$1.60/1M output).

\paragraph{\#Tokens.} The total tokens (input + output) consumed per question, reflecting the overall computational demand.

\paragraph{Cost-of-pass (\textcent).} The expected monetary cost to produce one correct answer, defined as ${Cost-of-pass} = C\,/\,{Acc}$ \citep{wang2025efficientagents}. By normalizing cost by accuracy, this metric penalizes both expensive systems and inaccurate ones, serving as our primary efficiency--effectiveness metric.

\subsection{Statistical Significance}
\label{appendix:significance}
We test whether GDP-RAG's improvements are statistically significant using a paired one-sided $t$-test, with a paired bootstrap of 10{,}000 resamples as a robustness check. Tests are run per question over the full filtered set ($n=1800$). Table~\ref{tab:significance} reports, for each metric, the mean per-question difference $\Delta$ (in points), the win rate (Win\%, the fraction of non-tied questions on which GDP-RAG scores higher), and the $p$-value. GDP-RAG significantly outperforms every external baseline on all three metrics ($p<0.001$). Against the strongest baseline PAR-RAG, the gains on EM and F1 are significant while SM saturates (Win\% $\approx 50$), so including SM in the Accuracy average makes our headline conservative rather than inflated. Among the ablations, the accuracy-oriented components (skeletal Thought, Review, Update) yield significant gains, whereas the efficiency-oriented planning ablations (GP, NP) are not significant, consistent with their role of reducing steps rather than improving accuracy.

\begin{table*}[h]
\centering
\small
\setlength{\tabcolsep}{4pt}
\resizebox{\textwidth}{!}{%
\begin{tabular}{l|ccc|ccc|ccc}
\toprule
\multirow{2}{*}{GDP-RAG vs.} & \multicolumn{3}{c|}{EM} & \multicolumn{3}{c|}{F1} & \multicolumn{3}{c}{SM} \\
\cmidrule(lr){2-4}\cmidrule(lr){5-7}\cmidrule(lr){8-10}
& $\Delta$ & Win\% & $p$ & $\Delta$ & Win\% & $p$ & $\Delta$ & Win\% & $p$ \\
\midrule
\multicolumn{10}{l}{\textit{Baselines}} \\
\midrule
Standard RAG & +20.61 & 85.2 & $<$.001 & {+22.58} & 80.0 & $<$.001 & {+26.72} & 88.4 & $<$.001 \\
IRCoT & {+17.72} & 82.0 & $<$.001 & {+17.97} & 75.9 & $<$.001 & {+19.44} & 81.1 & $<$.001 \\
KnowTrace & {+11.72} & 74.0 & $<$.001 & {+9.91} & 69.1 & $<$.001 & {+6.89} & 65.2 & $<$.001 \\
Godbole et al. & {+16.67} & 83.5 & $<$.001 & {+16.48} & 73.2 & $<$.001 & {+11.50} & 74.0 & $<$.001 \\
PAR-RAG & {+5.72} & 65.7 & $<$.001 & {+3.11} & 58.7 & $<$.001 & +0.06 & 50.2 & .48 \\
\midrule
\multicolumn{10}{l}{\textit{Planning-stage ablations}} \\
\midrule
GDP-RAG$_{GP}$ & +0.33 & 51.4 & .34 & +0.03 & 51.1 & .49 & $-$0.89 & 46.7 & .85 \\
GDP-RAG$_{NP}$ & +0.78 & 52.8 & .19 & +0.77 & 51.3 & .17 & +0.22 & 50.8 & .40 \\
GDP-RAG$_{RP}$ & +1.72 & 56.7 & .02 & +1.34 & 53.0 & .04 & +0.89 & 53.1 & .16 \\
GDP-RAG$_{NT}$ & {+3.89} & 63.4 & $<$.001 & {+4.20} & 61.0 & $<$.001 & {+4.95} & 65.1 & $<$.001 \\
\midrule
\multicolumn{10}{l}{\textit{Execution-stage ablations}} \\
\midrule
GDP-RAG$_{NR}$ & {+1.94} & 58.7 & .007 & {+2.13} & 56.1 & .001 & +1.17 & 54.8 & .08 \\
GDP-RAG$_{NU}$ & +1.44 & 56.4 & .03 & {+1.62} & 55.7 & .01 & +1.11 & 54.5 & .09 \\
\bottomrule
\end{tabular}
}
\caption{Statistical significance of GDP-RAG's per-question improvements over each baseline and ablation, on EM, F1, and SM. $\Delta$: mean score difference (points); Win\%: fraction of non-tied questions GDP-RAG wins; $p$: one-sided paired $t$-test (the bootstrap agrees throughout).}
\label{tab:significance}
\end{table*}

\subsection{Robustness to Model Scale (GPT-4.1-nano)}
\label{appendix:model_scale}
To test whether our findings depend on a particular backbone, we rerun every method with the weaker GPT-4.1-nano. Table~\ref{tab:nano-macro} reports macro accuracy and cost at both scales. Every method loses accuracy at the smaller scale, confirming the benchmark remains discriminative. GDP-RAG leads on accuracy at the mini scale; at the nano scale its accuracy lead narrows to a close second behind PAR-RAG (40.00 vs.\ 41.38), though it stays cheaper than PAR-RAG (0.07 vs.\ 0.09\textcent{} per query).

\begin{table}[h]
\centering
\small
\setlength{\tabcolsep}{5pt}
\begin{tabular}{l cc cc}
\toprule
\multirow{2}{*}{Method} & \multicolumn{2}{c}{GPT-4.1-mini} & \multicolumn{2}{c}{GPT-4.1-nano} \\
\cmidrule(lr){2-3}\cmidrule(lr){4-5}
& Acc\%$\uparrow$ & Cost\,\textcent{}$\downarrow$ & Acc\%$\uparrow$ & Cost\,\textcent{}$\downarrow$ \\
\midrule
Standard RAG & 37.86 & 0.04 & 24.11 & 0.01 \\
IRCoT & 44.56 & 0.06 & 35.78 & 0.01 \\
Godbole et al. & 46.28 & 0.45 & 32.84 & 0.11 \\
KnowTrace & 51.65 & 0.61 & 32.50 & 0.20 \\
PAR-RAG & 58.20 & 0.34 & \textbf{41.38} & 0.09 \\
\midrule
GDP-RAG & \textbf{60.63} & 0.27 & 40.00 & 0.07 \\
GDP-RAG$_{NR}$ & 59.41 & 0.21 & 37.79 & 0.06 \\
\bottomrule
\end{tabular}
\caption{Robustness to model scale. Macro accuracy and cost per query under GPT-4.1-mini and GPT-4.1-nano, averaged over the three datasets. \textbf{Bold} = best accuracy at each scale. Every method loses accuracy at the smaller scale.}
\label{tab:nano-macro}
\end{table}



To locate where the nano gap arises, Table~\ref{tab:nano-persplit} compares GDP-RAG against PAR-RAG, the strongest baseline, on every split at both scales. GDP-RAG matches or beats PAR-RAG on six of the eight nano splits. The two substantive deficits, 2WikiMultiHopQA 4-hop and MuSiQue 4-hop, are precisely the splits where GDP-RAG most outperforms PAR-RAG at the mini scale (+6.87 and +4.59). This reflects a tradeoff inherent to compression: GDP-RAG distills the plan to the missing facts and leaves the final compositional step to the answering model, which a capable model completes but a tiny one does not, whereas PAR-RAG's longer uncompressed trajectory carries spare steps that absorb a weak model's mistakes.

\begin{table}[h]
\centering
\small
\setlength{\tabcolsep}{4pt}
\begin{tabular}{l ccc ccc}
\toprule
\multirow{2}{*}{Split} & \multicolumn{3}{c}{GPT-4.1-mini} & \multicolumn{3}{c}{GPT-4.1-nano} \\
\cmidrule(lr){2-4}\cmidrule(lr){5-7}
& PAR-RAG & GDP-RAG & $\Delta$ & PAR-RAG & GDP-RAG & $\Delta$ \\
\midrule
HotpotQA 2-hop & 62.85 & 63.11 & +0.26 & 54.82 & 57.12 & +2.30 \\
HotpotQA 3-hop & 60.16 & 65.20 & +5.04 & 48.11 & 48.44 & +0.33 \\
HotpotQA 4-hop & 65.35 & 65.93 & +0.58 & 49.97 & 49.50 & $-$0.47 \\
2Wiki 2-hop & 63.34 & 63.51 & +0.17 & 54.72 & 55.40 & +0.69 \\
\rowcolor{blue!8} 2Wiki 4-hop & 74.87 & 81.74 & +6.87 & 32.60 & 19.72 & $-$12.88 \\
MuSiQue 2-hop & 57.40 & 54.52 & $-$2.88 & 39.79 & 41.24 & +1.45 \\
MuSiQue 3-hop & 39.44 & 43.24 & +3.80 & 21.18 & 29.71 & +8.53 \\
\rowcolor{blue!8} MuSiQue 4-hop & 31.25 & 35.84 & +4.59 & 27.56 & 21.33 & $-$6.23 \\
\bottomrule
\end{tabular}
\caption{Per-split accuracy (\%) of GDP-RAG vs.\ PAR-RAG at both model scales. $\Delta=$ GDP-RAG $-$ PAR-RAG. GDP-RAG matches or beats PAR-RAG on six of eight nano splits; the two deficits (2Wiki 4-hop, MuSiQue 4-hop) are the deepest-hop splits where GDP-RAG most outperforms PAR-RAG at mini.}
\label{tab:nano-persplit}
\end{table}

\subsection{Carry-forward Control: Delta Planning vs.\ Evidence Access}
\label{appendix:carryforward}
A natural question is whether GDP-RAG's gains come from carrying preliminary evidence forward rather than from delta planning itself. To isolate this, we give the Direct Planning baseline (PAR-RAG) the same preliminary passages $D_{rel}$, while leaving its planner unchanged---Direct Planning, with no gap-conditioning and no skeletal Thought. $D_{rel}$ is injected as side context either only into the final Answer module, or into every execution step (Act, Review, Update, and Answer); the retriever, model, and decoding all match the main experiments. As Table~\ref{tab:carryforward} shows, both variants fall below PAR-RAG alone while costing more, and neither approaches GDP-RAG. Simply carrying $D_{rel}$ forward, without delta planning to distill it into the skeletal Thought, adds noise rather than signal: the benefit lies in \emph{how} the evidence is used, not in whether it is available.

\begin{table*}[t]
\centering
\small
\setlength{\tabcolsep}{3.5pt}
\resizebox{\textwidth}{!}{%
\begin{tabular}{l|cccc|cccc|cccc}
\toprule
\multirow{2}{*}{Method} & \multicolumn{4}{c|}{cost-of-pass$\downarrow$} & \multicolumn{4}{c|}{Acc./\%$\uparrow$} & \multicolumn{4}{c}{Cost (\textcent{})$\downarrow$} \\
\cmidrule(lr){2-5}\cmidrule(lr){6-9}\cmidrule(lr){10-13}
& all & l1 & l2 & l3 & all & l1 & l2 & l3 & all & l1 & l2 & l3 \\
\midrule
PAR-RAG & 0.65 & 0.39 & 0.43 & 1.14 & 58.20 & 62.79 & 69.10 & 42.70 & 0.34 & 0.24 & 0.30 & 0.49 \\
PAR-RAG + $D_{rel}$ (answer only) & 0.73 & 0.41 & 0.46 & 1.31 & 57.32 & 62.38 & 68.74 & 40.85 & 0.37 & 0.26 & 0.32 & 0.54 \\
PAR-RAG + $D_{rel}$ (every step) & 0.92 & 0.46 & 0.54 & 1.77 & 55.22 & 64.31 & 61.07 & 40.28 & 0.45 & 0.30 & 0.33 & 0.71 \\
\midrule
GDP-RAG & \textbf{0.51} & \textbf{0.25} & \textbf{0.30} & \textbf{0.98} & \textbf{60.63} & \textbf{64.75} & \textbf{72.62} & \textbf{44.53} & \textbf{0.27} & \textbf{0.16} & \textbf{0.22} & \textbf{0.44} \\
\bottomrule
\end{tabular}
}
\caption{Carry-forward control. Direct Planning (PAR-RAG) given the same preliminary retrieval $D_{rel}$, injected at the final answer only or at every execution step, with all other settings matched to the main experiments. Adding $D_{rel}$ without delta planning does not reach GDP-RAG and even falls below PAR-RAG alone. \textbf{Bold} = best.}
\label{tab:carryforward}
\end{table*}

\subsection{Grounded Delta Planning Advantages Analysis}
\label{sec:grounded_planning_analysis}
To quantify the efficacy of Grounded Delta Planning, we perform an LLM-based alignment. We feed the reasoning steps generated by both Direct Planning and Grounded Delta Planning alongside the Gold Documents into an LLM to identify valid mappings. Based on this alignment, the metrics in Figure~\ref{fig:planning_stats} are defined as follows:

\paragraph{Reasoning-step Compression.} Under the same alignment criteria, we measure valid reasoning steps by counting generated queries that match gold documents. The reduction in this count from Direct to Grounded Planning represents necessary steps saved because the planner recognizes information already available in the retrieved context.

\begin{table}[h]
\centering
\small
\setlength{\tabcolsep}{6pt}
\begin{tabular}{l cccc}
\toprule
Dataset & 2-hop & 3-hop & 4-hop & Avg \\
\midrule
HotpotQA & 0.86 & 0.77 & 0.71 & 0.78 \\
2WikiMultiHopQA & 0.66 & --- & 0.50 & 0.58 \\
MuSiQue & 0.70 & 0.44 & 0.24 & 0.46 \\
\bottomrule
\end{tabular}
\caption{Preliminary retrieval recall@10 by dataset and hop (macro average 0.61). Recall declines with hop count and closely tracks the 61.0\% Reasoning-step Compression (Section~\ref{sec:effectiveness}).}
\label{tab:recall-hop}
\end{table}

\paragraph{Over-planning Pruning.} Under the same alignment criteria, we quantify redundant planning by counting steps that fail to align with any gold document. The difference between the two settings represents redundant queries eliminated through grounding, as the planner avoids generating steps for irrelevant or already-covered information.

\paragraph{Under-planning Refinement.} This metric quantifies missed information by counting gold documents not covered by the system. While Direct Planning is evaluated against generated reasoning steps alone, Grounded Delta Planning is measured against the union of preliminary retrieval ($D_{rel}$) and the generated plan. The reduction in missing documents represents information recovered by incorporating initial retrieval into the planning process, ensuring that essential details overlooked by the planner are still accounted for.

\paragraph{Step-matching Validation}
We validated LLM accuracy via human review on a 10\% sample (Table \ref{tab:multihop_results}). Four researchers, each holding at least a bachelor's degree, conducted this audit to ensure the expertise required for evaluating complex multi-hop reasoning. For both settings, we verified whether each step-matching output correctly matched a gold document or was accurately identified as redundant.

\begin{table}[h]
\centering
\small
\begin{tabular}{lrrrr}
\midrule
\textbf{Dataset} & \textbf{Total} & \textbf{Correct} & \textbf{Acc (\%)} \\
\midrule
HotpotQA 2hop & 84  & 83  & 98.81 \\
HotpotQA 3hop & 77  & 60  & 77.92 \\
HotpotQA 4hop & 87  & 78  & 89.66 \\
MuSiQue 2hop  & 83  & 68  & 81.93 \\
MuSiQue 3hop  & 103 & 77  & 74.76 \\
MuSiQue 4hop  & 145 & 125 & 86.21 \\
2Wiki 2hop    & 118 & 101 & 85.59 \\
2Wiki 4hop    & 219 & 159 & 72.60 \\
\midrule
\textbf{Total} & \textbf{916} & \textbf{751} & \textbf{81.99} \\
\midrule
\end{tabular}
\caption{LLM Step-matching Accuracy}
\label{tab:multihop_results}
\end{table}

\subsection{Plan-level Analysis under a Frontier Planner}
\label{appendix:plan_scale}
A natural question is whether a stronger planner already produces concise plans on its own, making Grounded Delta Planning unnecessary. To test this, we repeat the plan-level analysis of Section~\ref{sec:effectiveness} with the frontier model GPT-5.5 as the planner, reporting the necessary and redundant step counts behind Reasoning-step Compression and Over-planning Pruning. As Table~\ref{tab:plan-scale} shows, a stronger Direct planner does write sub-questions that align to more gold passages (necessary steps rise from 3{,}561 to 4{,}525), yet it continues to over-plan (2{,}609 redundant steps). Grounded Delta Planning still compresses the necessary steps to 1{,}673 and prunes the redundant steps to 506. Both advantages are therefore structural: compression comes from preliminary retrieval resolving facts before planning, and over-planning is a property of planning without sight of the retrieved context, neither of which a stronger planner removes on its own. For this study the GPT-5.5 plans are generated in a planning-only setting, using the same preliminary retrieval $D_{rel}$ and the same fixed judge (GPT-4.1-mini) as Figure~\ref{fig:planning_stats}, so the comparison isolates the effect of planner strength. The GPT-4.1-mini column therefore reproduces the values of that figure.

\begin{table}[h]
\centering
\small
\setlength{\tabcolsep}{5pt}
\begin{tabular}{l cc cc}
\toprule
\multirow{2}{*}{Step type} & \multicolumn{2}{c}{Direct Planning} & \multicolumn{2}{c}{Grounded Delta (Ours)} \\
\cmidrule(lr){2-3}\cmidrule(lr){4-5}
& GPT-4.1-mini & GPT-5.5 & GPT-4.1-mini & GPT-5.5 \\
\midrule
Necessary  & 3{,}561 & 4{,}525 & 1{,}388 & 1{,}673 \\
Redundant & 2{,}823 & 2{,}609 & 650 & 506 \\
\bottomrule
\end{tabular}
\caption{Plan-level analysis under a frontier planner. Necessary and Redundant step counts from Direct Planning and Grounded Delta Planning under GPT-4.1-mini and GPT-5.5. The GPT-4.1-mini columns are the values of Figure~\ref{fig:planning_stats}.}
\label{tab:plan-scale}
\end{table}

\subsection{Qualitative Case Analysis}
\label{appendix:case-study}
Tables~\ref{table:case-compression}--\ref{table:case-refinement} illustrate the three structural advantages of Grounded Delta Planning on concrete examples. In each, \textbf{GD} denotes a Gold Document, \textbf{S} a Subquery, colors indicate supporting context, and {[NULL]} marks a redundant step matched to no gold document.

\paragraph{Reasoning-step Compression.}
Direct Planning issues multiple subqueries to resolve intermediate facts. In the first example, it generates two steps: one to identify the academy and another to locate it. Grounded Delta Planning, having already retrieved both gold documents (GD1, GD2) during preliminary retrieval, compresses these into a single step that directly asks for the location (Table~\ref{table:case-compression}).

\paragraph{Over-planning Pruning.}
Direct Planning generates redundant steps that do not align with any gold document. In the second example, it produces three steps: two to retrieve the founding years and a third comparison step (marked [NULL]) whose inputs are already available. Grounded Delta Planning identifies that both founding years are already in $D_{rel}$ and generates only a single comparison step, eliminating the redundant action (Table~\ref{table:case-pruning}).

\paragraph{Under-planning Refinement.}
Direct Planning decomposes the query into narrow, targeted subqueries that may miss relevant documents. In the third example, Direct Planning generates three steps but only covers two of the three gold documents (GD1, GD2), missing GD3 entirely. Because Grounded Delta Planning retrieves using the original query's broader semantics, it recovers all three gold documents (GD1, GD2, GD3) during preliminary retrieval, capturing dependencies that the targeted subqueries would miss (Table~\ref{table:case-refinement}).

\begin{table}[h]
\centering
\scriptsize
\renewcommand{\arraystretch}{1.3}
\begin{tabular}{@{}p{\columnwidth}@{}}
\toprule
\textbf{Question:} Where is the academy, for which Joseph D. Stewart was appointed Superintendent, located? \\
\midrule
\textbf{Gold Documents:} \\
\textcolor{Blue}{\textbf{GD1}}: Joseph D. Stewart \ldots\ was appointed as Superintendent of the United States Merchant Marine Academy \ldots \\
\textcolor{OliveGreen}{\textbf{GD2}}: The United States Merchant Marine Academy \ldots\ located in Kings Point, New York. \\
\midrule
\textbf{Direct Planning:} \\
S1: For which \textcolor{Blue}{\textbf{academy was Joseph D. Stewart appointed Superintendent [GD1]}}? \\
S2: \textcolor{OliveGreen}{\textbf{Where is the academy}}, for which Joseph D. Stewart was appointed Superintendent, \textcolor{OliveGreen}{\textbf{located [GD2]}}? \\
\midrule
\textbf{Grounded Delta Planning:} \\
\textit{Retrieved Context}: \textcolor{Blue}{\textbf{[GD1]}}, \textcolor{OliveGreen}{\textbf{[GD2]}} \\
S1: Where is the United States Merchant Marine Academy located? \\
\bottomrule
\end{tabular}
\caption{Reasoning-step Compression. Direct Planning uses two steps to identify and then locate the academy. Grounded Delta Planning, already holding GD1 and GD2 from preliminary retrieval, compresses them into a single step.}
\label{table:case-compression}
\end{table}

\begin{table}[h]
\centering
\scriptsize
\renewcommand{\arraystretch}{1.3}
\begin{tabular}{@{}p{\columnwidth}@{}}
\toprule
\textbf{Question:} Was Vanderbilt University or Emory University founded first? \\
\midrule
\textbf{Gold Documents:} \\
\textcolor{Blue}{\textbf{GD1}}: Vanderbilt University -- Founded in 1873, \ldots \\
\textcolor{OliveGreen}{\textbf{GD2}}: Emory University \ldots\ was founded as Emory College in 1836 in Oxford \ldots \\
\midrule
\textbf{Direct Planning:} \\
S1: What year was \textcolor{Blue}{\textbf{Vanderbilt University founded [GD1]}}? \\
S2: What year was \textcolor{OliveGreen}{\textbf{Emory University founded [GD2]}}? \\
S3: Which was founded first \ldots\ {[NULL]} \\
\midrule
\textbf{Grounded Delta Planning:} \\
\textit{Retrieved Context}: \textcolor{Blue}{\textbf{[GD1]}}, \textcolor{OliveGreen}{\textbf{[GD2]}} \\
S1: Based on the founding years provided, which university was founded first: Vanderbilt University or Emory University? \\
\bottomrule
\end{tabular}
\caption{Over-planning Pruning. Direct Planning adds a redundant comparison step whose inputs are already retrieved. Grounded Delta Planning prunes it to a single comparison.}
\label{table:case-pruning}
\end{table}

\begin{table}[h]
\centering
\scriptsize
\renewcommand{\arraystretch}{1.3}
\begin{tabular}{@{}p{\columnwidth}@{}}
\toprule
\textbf{Question:} What is another name for a forest which has the hamlet of Oakenclough on the edge of it? \\
\midrule
\textbf{Gold Documents:} \\
\textcolor{Blue}{\textbf{GD1}}: Oakenclough is an English hamlet located on the edge of the Forest of Bowland in Lancashire. \\
\textcolor{OliveGreen}{\textbf{GD2}}: The Forest of Bowland, also known as the Bowland Fells, \ldots \\
\textcolor{RedOrange}{\textbf{GD3}}: It was once described as the ``Switzerland of England''. \\
\midrule
\textbf{Direct Planning:} \\
S1: Where is the \textcolor{Blue}{\textbf{hamlet of Oakenclough located [GD1]}}? \\
S2: What forest \textcolor{OliveGreen}{\textbf{lies on the edge of the hamlet Oakenclough [GD2]}}? \\
S3: What is \textcolor{OliveGreen}{\textbf{another name for the forest [GD2]}} that has Oakenclough on its edge? \\
\midrule
\textbf{Grounded Delta Planning:} \\
\textit{Retrieved Context}: \textcolor{Blue}{\textbf{[GD1]}}, \textcolor{OliveGreen}{\textbf{[GD2]}}, \textcolor{RedOrange}{\textbf{[GD3]}} \\
S1: What is another name for the forest which has the hamlet of Oakenclough on the edge of it? \\
\bottomrule
\end{tabular}
\caption{Under-planning Refinement. Direct Planning's targeted subqueries miss GD3. Grounded Delta Planning's broader preliminary retrieval recovers all three gold documents.}
\label{table:case-refinement}
\end{table}

\subsection{Workflow}
Algorithm~\ref{alg:main} summarizes the full GDP-RAG procedure, following the three phases of Section~\ref{sec:GDP-RAG-framework}.
In the {Planning phase}, the system performs preliminary retrieval to obtain query-relevant context $D_{rel}$ and calls the gap-conditioned planner to produce a skeletal trajectory $\mathcal{T}_{skel}$ of $(\theta_t, q_t)$ pairs that target only the information delta.
The {Trajectory Generation phase} then grounds each planned step through the Act--Review--Update cycle: Act retrieves documents $d_t$ for $q_t$ and drafts a provisional answer $\hat{a}_t$ from the current Trajectory Memory $\mathcal{M}_{t-1}$; Review issues a second, answer-conditioned retrieval $d_t^*$ and cross-verifies $\hat{a}_t$ into a validated answer $a_t$; and Update commits the grounded tuple to $\mathcal{M}_t$ and refines the next subquery $q_{t+1}$ from the accumulated evidence. Providing $\mathcal{M}_{t-1}$ to all three modules keeps the steps consistent.
Once all $n$ steps are grounded, the {Answering phase} synthesizes the final answer $A$ from the completed trajectory $\mathcal{T} = \mathcal{M}_n$.

\begin{algorithm}[h]
\caption{Grounded Delta Planning RAG}
\label{alg:main}
\begin{algorithmic} 
\Require Question $Q$, Document Collection $\mathcal{D}$, Trajectory Memory $\mathcal{M}_{0}$
\Ensure Final Answer $A$
    \State $D_{rel} = {Retrieve}(Q, \mathcal{D})$ \Comment{Preliminary Retrieval}
    \State $\mathcal{T}_{skel} = {Plan}(Q, D_{rel})$ \Comment{Gap-aware Decomposition}
    \For{$t = 1, \dots, n$}
        \State $(\theta_t, q_t) \gets \mathcal{T}_{skel}$
        \State $d_t = {Retrieve}(q_t, \mathcal{D})$
        \State $\hat{a}_t = {Act}(q_t, d_t, \mathcal{M}_{t-1})$ \Comment{Provisional answer}
        \State $d_t^* = {Retrieve}(q_t + \hat{a}_t, \mathcal{D})$
        \State $a_t = {Review}(q_t, \hat{a}_t, d_t^*, \mathcal{M}_{t-1})$ \Comment{Cross-verify}
        \State $(\mathcal{M}_t, q_{t+1}) = {Update}(q_t, a_t, d_t \cup d_t^*, \mathcal{M}_{t-1})$
    \EndFor
    \State $\mathcal{T} = \mathcal{M}_n$
    \State $A = {Answer}(Q, \mathcal{T})$
\end{algorithmic}
\end{algorithm}

\end{document}